\documentclass[sigconf]{acmart}
\AtBeginDocument{%
  }

\setcopyright{acmlicensed}
\copyrightyear{2026}
\acmYear{2026}
\acmDOI{XXXXXXX.XXXXXXX}
\acmConference[ICAIL 2026]{21\textsuperscript{st} International Conference on Artificial Intelligence and Law}{June 08--12, 2026}{Singapore}
\acmISBN{XXX-X-XXXX-XXXX-X/XXXX/XX}

\acmSubmissionID{154}

\usepackage{graphicx} 
\usepackage{tikz}
\usetikzlibrary{arrows.meta,positioning,fit,calc,shapes.geometric,backgrounds}
\usepackage{enumitem}

\begin{document}

\title{LegalCheck: Retrieval- and Context-Augmented Generation for Drafting Municipal Legal Advice Letters}

\author{Virgill van der Meer}

\orcid{1234-5678-9012}
\affiliation{%
  \institution{Municipality of Amsterdam}
  \city{Amsterdam}
  \country{Netherlands}
}
\email{v.van.der.meer@amsterdam.nl}

\author{Julien Rossi}
\affiliation{%
  \institution{University of Amsterdam}
  \city{Amsterdam}
  \country{Netherlands}
}
\email{j.rossi@uva.nl}

\renewcommand{\shortauthors}{van der Meer and Rossi}

\begin{abstract}
Public-sector legal departments in the Netherlands face acute staff shortages, increased case volumes, and increased pressure to meet regulatory compliance. This paper presents LegalCheck, a novel system that addresses these challenges by automating the drafting of objection response letters through a combination of Retrieval-Augmented Generation (RAG) and Context-Augmented Generation (CAG). Using a large language model (LLM) alongside curated legal knowledge bases, LegalCheck performs retrieval of relevant laws and precedents, and uses controlled prompting to incorporate both external knowledge and case-specific details into a coherent draft. An expert-in-the-loop review ensures that each generated letter is legally sound and contextually appropriate. In a real-world deployment within the Municipality of Amsterdam, LegalCheck produced near-final advice letters in minutes rather than hours, while maintaining high legal consistency and factual accuracy. The output is based on actual regulations and prior cases, providing explainable outputs that captured the vast majority of required legal reasoning (often 80\% to 100\% of essential content). Legal professionals found that the system reduced their workload and ensured a consistent application of legal standards, without replacing human judgment. These results demonstrate substantial efficiency gains, improved legal consistency, and positive user acceptance. More broadly, this work illustrates how responsible AI can be deployed in the legal domain by augmenting LLMs with domain knowledge and governance mechanisms.
\end{abstract}

\begin{teaserfigure}
  \includegraphics[width=\textwidth]{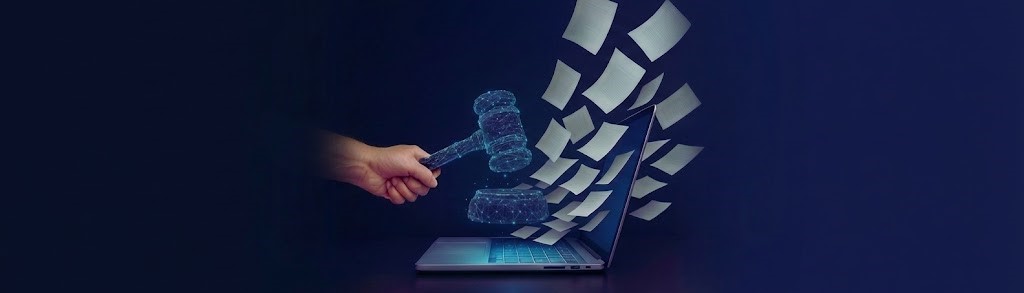}
  \Description{The human hand on the (digital) gavel shows that hybrid human–AI intelligence is meant to augment, not replace, the judgment of legal professionals.}
  \label{fig:teaser}
\end{teaserfigure}

\maketitle

\section{Introduction}
Municipal legal departments face increasing caseloads and persistent staff shortages, creating pressure to deliver more with fewer resources \cite{VNG2024Pilot} \cite{pwc2023dutchjobsgenai}. In the City of Amsterdam’s Legal Department, the workload associated with handling formal citizen objections\footnote{In Dutch administrative law, the objection procedure (\emph{bezwaar}) is a statutory internal review mechanism. Before going to court, citizens can ask the administrative body to reconsider its decision, with a right to be heard at low threshold. It plays a key role in legal protection, functioning both as a learning mechanism for the administration and as a filter that prevents unnecessary court cases.} in administrative and environmental law has increased by 35\% over the last five years. This environment has driven a transformation to AI-based and data-driven workflows \cite{faraj2018learningalgorithm}. A centralized digital case library (\textit{Zaakbibliotheek}) was established, consolidating ~55,000 past cases (objections and decisions) into an indexed knowledge base. Building on this data foundation, the department began piloting AI tools to alleviate routine tasks and improve efficiency. In particular, several supportive applications were developed: \textit{Kern van de Zaak}, an AI-based summarization tool that extracts the core issues from legal documents to enable faster research; \textit{Juris-Match}, a semantic case search engine for finding conceptually similar precedents; \textit{Spraakzaak}, a speech-to-text transcription and summarization system for hearings; and the \textit{KOG-i-fier}, a classifier that triages incoming objections by labeling straightforward (clearly unfounded) cases using generative AI. Each tool addresses a specific pain point – from legal research and case intake to transcript handling – illustrating how AI can unlock the value of legal data and automate the clerical workload in the public sector.

The flagship of this AI-driven transformation is \textit{LegalCheck}, a system that drafts preliminary legal advice letters for citizen objection cases. Legal advice letters are complex, high-stakes documents: they must accurately cite laws and prior rulings, align with case facts, and maintain a professional yet citizen-friendly tone. Manual drafting of these letters is both time-intensive and cognitively demanding, as it requires meticulous legal reasoning, contributing to backlogs under heavy caseloads. The core idea of LegalCheck follows a human-in-the-loop paradigm: the model produces a grounded draft, while a legal expert remains responsible for reviewing, correcting, and approving the final letter. This approach acknowledges government guidelines that speed gains from AI must not come at the expense of due diligence and human judgment \cite{europeancommission2019trustworthyai} \cite{amsterdam2024visionai}. Developing “augmentation rather than automation,” LegalCheck is positioned as an assistant to legal staff, with the aim of improving productivity, substantive quality, and consistency in decision-writing while maintaining accountability and trust \cite{Davenport2018RealWorld} \cite{bommasani2021foundationmodels}.

This research contributes to the growing literature on the topic of RAG and expands it by presenting the first applied study (to our knowledge) of using RAG for legal document drafting in a real public-sector setting. This research bridges recent advances in AI language models with the practical needs and constraints of government legal practice. Unlike previous legal NLP research that mostly addressed classification or outcome prediction \cite{Katz2017SCOTUS}, or recent studies of AI assistance in law \cite{chienkim2024legalaid} \cite{wrzesniowska2023canaimakeacase}, LegalCheck demonstrates \textit{in vivo} how generative AI can produce high-quality legal reasoning \cite{Schwarcz2025AIPowered} on scale when coupled with a domain-specific knowledge base and human oversight \cite{Aletras2016ECtHR} \cite{thomsonreuters2025lessdrudge}.

The system introduces a novel Context-Augmented Generation (CAG) pipeline layered on top of standard RAG. In addition to retrieving relevant documents to ground the model’s output, LegalCheck integrates case-specific input beyond the existing knowledge base. Specifically, the jurist can upload key documents related to the case – such as the citizen’s objection letter and the official report of the enforcement officer –, specify the intended dictum (uphold or reject), and optionally provide free-text “steering advice” that reflects expert knowledge or instructions not captured elsewhere. After LegalCheck produces an initial draft, the human reviewer can provide contextual feedback by highlighting omissions or suggesting revisions. The AI then performs a refinement pass, regenerating the letter to incorporate the expert’s input.

This multi-stage CAG approach uses the substantive content of the new case, expert guidance and refinement, and the structure of previous similar cases to improve the precision and completeness of the draft. We implement CAG through multiple LLM prompt stages. In the initial prompt stage, the model is provided with the uploaded case documentation along with any jurist annotations and generates a draft that integrates all relevant information. In the refinement stage, the model is prompted with additional annotations on this generated draft and returns a revised version that incorporates the requested corrections. 

Our contributions are twofold: (1) an integrated RAG+CAG architecture that produces quasi-final legal advice letters with proper citations and a formal, accessible tone suitable for communication with citizens and (2) an empirical evaluation of the impact of this system on the efficiency of the drafting, output quality, and professional workflows in a municipal legal department.

We position this work in the growing academic discourse on “AI-powered lawyering,” providing empirical evidence that retrieval-based text generation can significantly improve the quality and speed of legal work \cite{Schwarcz2025AIPowered}. In addition, we address the organizational and ethical context: the project is deployed in a government setting subject to strict privacy, security, and transparency requirements. Thus, the research contributes case-based insights on responsible AI adoption in the legal domain, aligning with calls for trustworthy AI in public services, including how to maintain legal precision, explainability, and human control when using large language models (LLMs) for official decision documents \cite{europeancommission2019trustworthyai}.

The remainder of this paper is structured as follows. Section 2 reviews related work on legal NLP, retrieval-augmented generation, and human–AI workflows in high-stakes legal settings. Section 3 describes the LegalCheck system architecture, including the RAG pipeline and the multi-stage CAG pipeline that integrates case-specific context before and after initial generation, and situates it relative to existing AI techniques in law. Section 4 details our methodology, including the datasets of waste fine objections and the removal and towing of bicycles, motor vehicles, and boats the semantic indexing and retrieval process, the prompt design, and our evaluation setup (metrics and expert-in-the-loop validation). Section 5 reports the results of both quantitative evaluations (output quality, speed gains) and qualitative feedback from legal professionals. We also discuss the legal reasoning of the system, the role of the human expert, and trust in the AI-assisted process. Section 6 concludes with implications for legal practice and future research directions, including the reproducibility and generalizability of our approach.

\section{Literature Review}

\subsection{Legal NLP and Early AI in Law}
Research on AI in the legal domain has traditionally focused on analysis and prediction tasks rather than text generation \cite{surden2019aioverview}. Many foundational legal NLP works applied machine learning to classify documents or predict case outcomes \cite{ashley2017ailegalanalytics}.

For example, Aletras et al. (2016) used textual features to predict decisions of the European Court of Human Rights \cite{Aletras2016ECtHR}, and Katz et al. (2017) built a model to forecast U.S. Supreme Court verdicts \cite{Katz2017SCOTUS}.
These approaches demonstrated that NLP can derive patterns from legal text for tasks such as prediction of judgment, aligning with broader efforts in legal analytics \cite{ashley2017ailegalanalytics}. However, they did not attempt to generate new legal documents. Only recently, with the advent of generative models, have researchers begun to explore AI for the drafting of legal text \cite{shao2025when}. Early explorations have either fine-tuned LLMs on relatively small legal drafting corpora to generate prototype documents, or used retrieval-augmented models for narrow legal question answering on case fragments rather than for end-to-end drafting of official legal letters \cite{lincheng2024legaldrafting,wiratunga2024cbrrag}.

In particular, Lin \& Cheng (2024) showed that a fine-tuned transformer could draft legal documents in a lab setting \cite{lincheng2024legaldrafting}, and Wiratunga et al. (2024) introduced a case-based retrieval approach to improve the answer of the LLM to legal questions \cite{wiratunga2024cbrrag}. These studies hint at the potential of generative AI in law but remain confined to experimental or narrowly-scoped scenarios.

\subsection{Retrieval-Augmentation and Human–AI Workflow}
A key challenge in legal text generation is ensuring accuracy and trustworthiness. Therefore, recent techniques combine LLMs with external knowledge. Retrieval-Augmented Generation (RAG)~\cite{Lewis2020RAG} grounds an LLM’s output in relevant documents and has been applied to knowledge-intensive tasks to reduce errors.
In the legal context, grounding is crucial to prevent hallucinations (fabrications of laws or facts) by the model. By retrieving actual statutes or precedents and feeding them into the prompt, AI’s suggestions can be verified and legally sound \cite{Lewis2020RAG} \cite{magesh2024hallucinationfree}.
Our approach adopts this strategy, drawing on a curated case library so that every generated argument can be traced to real source material. Equally important is the role of human expertise. Experts stress that in high-stakes domains such as law, AI systems should augment rather than replace human judgment \cite{amershi2019guidelines} \cite{europeancommission2019trustworthyai}.
Following this principle, we implement a human-in-the-loop review: the AI system generates a draft advice letter, and a legal professional rigorously checks and edits it. This design aligns with previous findings that combining human expertise with AI can yield better results than either alone, provided that the human supervises and corrects the AI output \cite{amershi2019guidelines} \cite{holstein2019improvingfairness}, and with recent work that conceptualizes meaningful human oversight as a layered allocation of agency between AI “proposers” and human “evaluators” \cite{zhu2025oversight}.
Recent experimental studies underscore the need for such oversight: GPT-3.5 could “make a case” in Dutch law comparable to a junior lawyer, but human review was still necessary for nuanced correctness~\cite{wrzesniowska2023canaimakeacase}, legal aid attorneys benefited from GPT-generated drafts for client letters, but only after verifying and refining the content~\cite{chienkim2024legalaid}.

\subsection{Novelty of Our Approach}

In light of the literature above, LegalCheck is, to our knowledge, the first system to deploy an LLM-based drafting assistant for official municipal legal letters in a real public-sector workflow. Previous works on AI-generated legal writing were proof-of-concept studies (e.g. a fine-tuned model writing a generic legal document \cite{lincheng2024legaldrafting}) or simulations (comparing AI and human writing on a sample task \cite{wrzesniowska2023canaimakeacase}), without integrating into everyday legal operations.
By contrast, our work integrates a GPT-4o-based pipeline with the actual processes of a municipal legal department, handling live citizen objection cases. This use case – drafting municipal objection response letters – has not been addressed in the literature so far. We fill this gap by demonstrating a complete hybrid human–AI workflow for decision drafting.

\section{System Architecture and Technical Approach}

\subsection{High-Level}

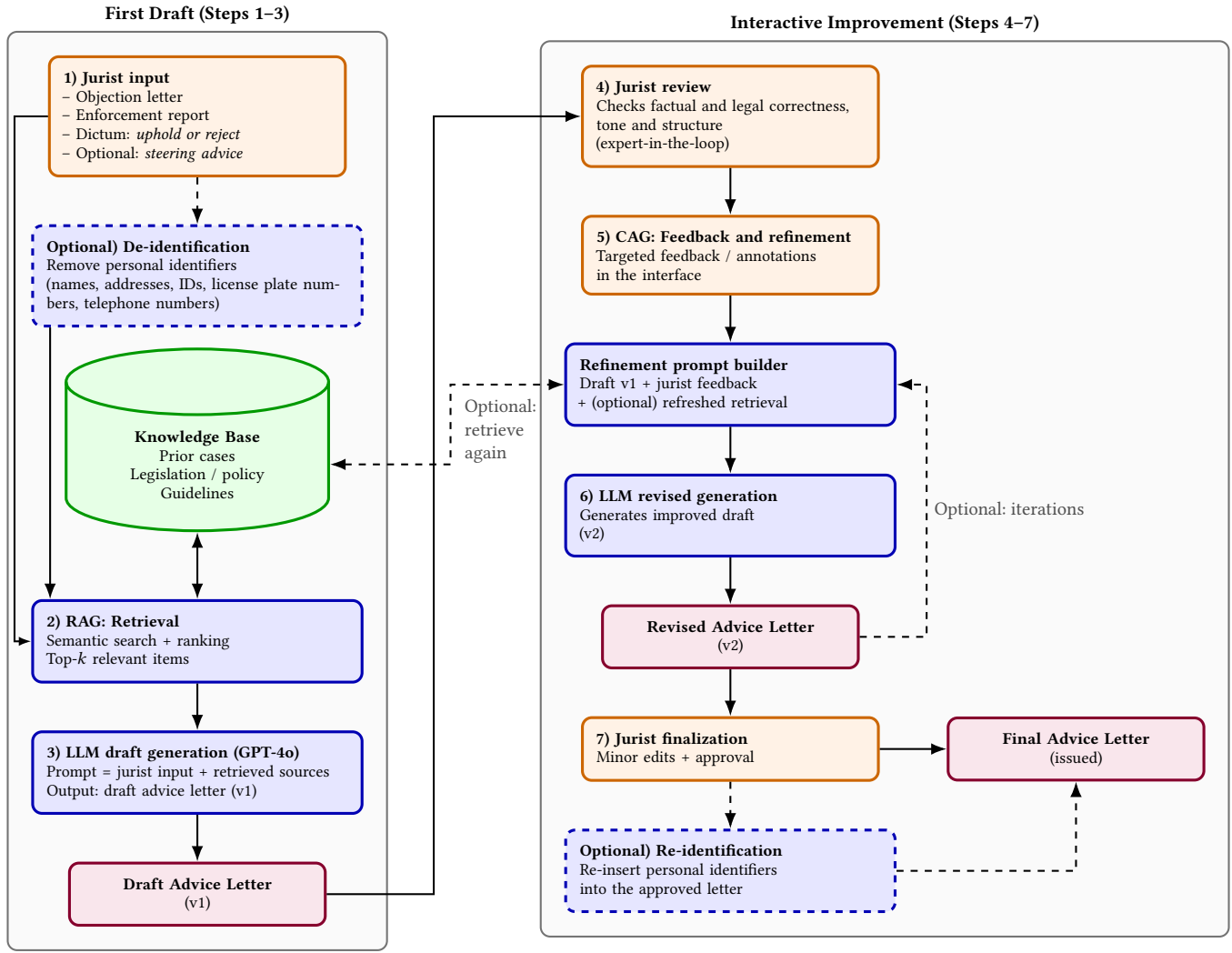
\begin{figure*}[t]
\centering
\resizebox{\textwidth}{!}{%
\begin{tikzpicture}[
  font=\footnotesize,
  node distance=8mm and 20mm,
  >=Latex,
  human/.style={
    draw=orange!80!black, very thick, rounded corners,
    fill=orange!10, align=left,
    inner xsep=6pt, inner ysep=6pt,
    text width=4.0cm
  },
  ai/.style={
    draw=blue!70!black, very thick, rounded corners,
    fill=blue!10, align=left,
    inner xsep=6pt, inner ysep=6pt,
    text width=4.5cm
  },
  data/.style={
    draw=green!60!black, very thick,
    fill=green!10, align=center,
    inner xsep=6pt, inner ysep=6pt,
    text width=3.5cm,
    cylinder, shape border rotate=90, aspect=0.25
  },
  output/.style={
    draw=purple!70!black, very thick, rounded corners,
    fill=purple!10, align=center,
    inner xsep=6pt, inner ysep=6pt,
    text width=3.4cm
  },
  groupbox/.style={
    draw=black!55, thick, rounded corners,
    fill=black!2, inner sep=10pt
  },
  arrow/.style={->, thick},
  dashedarrow/.style={->, thick, dashed}
]

\node[human] (s1) {\textbf{1) Jurist input}\\
-- Objection letter\\
-- Enforcement report\\
-- Dictum: \emph{uphold or reject} \\
-- Optional: \emph{steering advice}};

\node[ai, dashed, below=7mm of s1] (s1b) {\textbf{Optional) De-identification}\\
Remove personal identifiers\\
(names, addresses, IDs, license plate numbers, telephone numbers)};

\node[data, below=3mm of s1b] (kb) {\textbf{Knowledge Base}\\
Prior cases\\
Legislation / policy\\
Guidelines};

\node[ai, below=10mm of kb] (s2) {\textbf{2) RAG: Retrieval}\\
Semantic search + ranking\\
Top-$k$ relevant items};

\node[ai, below=7mm of s2] (s3) {\textbf{3) LLM draft generation (GPT-4o)}\\
Prompt = jurist input + retrieved sources\\
Output: draft advice letter (v1)};

\node[output, below=7mm of s3] (v1) {\textbf{Draft Advice Letter}\\(v1)};

\draw[dashedarrow] (s1.south) -- (s1b.north);
\draw[arrow] (s1b.south)+(-22mm,0) -- ($(s2.north)+(-22mm,0)$);
\draw[<->, thick] (s2.north) -- (kb.south);
\draw[arrow] (s2.south) -- (s3.north);
\draw[arrow] (s3.south) -- (v1.north);
\draw[-{Latex[scale=0.8]}, thick] (s1.west) -- ++(-5mm,0) |- (s2.west);

\node[human, right=35mm of s1] (s4) {\textbf{4) Jurist review}\\
Checks factual and legal correctness,\\
tone and structure\\
(expert-in-the-loop)};

\node[human, below=7mm of s4] (s5) {\textbf{5) CAG: Feedback and refinement}\\
Targeted feedback / annotations\\
in the interface};

\node[ai, below=7mm of s5] (s5b) {\textbf{Refinement prompt builder}\\
Draft v1 + jurist feedback\\
+ (optional) refreshed retrieval};

\node[ai, below=7mm of s5b] (s6) {\textbf{6) LLM revised generation}\\
Generates improved draft\\
(v2)};

\node[output, below=7mm of s6] (v2) {\textbf{Revised Advice Letter}\\(v2)};

\node[human, below=7mm of v2] (s7) {\textbf{7) Jurist finalization}\\
Minor edits + approval};

\node[ai, dashed, below=7mm of s7] (s8) {\textbf{Optional) Re-identification}\\
Re-insert personal identifiers\\
into the approved letter};

\node[output, right=10mm of s7] (v3) {\textbf{Final Advice Letter}\\(issued)};

\draw[arrow] (s4.south) -- (s5.north);
\draw[arrow] (s5.south) -- (s5b.north);
\draw[arrow] (s5b.south) -- (s6.north);
\draw[arrow] (s6.south) -- (v2.north);
\draw[arrow] (v2.south) -- (s7.north);
\draw[dashedarrow] (s7.south) -- (s8.north);
\draw[arrow] (s7.east) -- (v3.west);
\draw[dashedarrow] (s8.east) -| (v3.south);

\draw[arrow] (v1.east) -- ++(16mm,0) |- (s4.west);

\draw[dashedarrow]
  (v2.east) -- ++(10mm,0) coordinate (v2turn)
  -- (v2turn |- s5b.east) node[midway, right, font=\small, text=black!70]{Optional: iterations}
  -- (s5b.east);
\draw[Latex-Latex, thick, dashed]
  (kb.east) -- ++(18mm,0) coordinate (kbturn) |- (s5b.west);
\node[right of=kbturn, font=\small, text=black!70, align=left, xshift=-0.5mm, yshift=5mm]{Optional:\\retrieve\\again};
\begin{scope}[on background layer]
  \node[groupbox, fit=(s1) (s1b) (kb) (s2) (s3) (v1),
        label={[font=\bfseries\small]above: First Draft (Steps 1--3)}] {};
  \node[groupbox, fit=(s4) (s5) (s5b) (s6) (v2) (s7) (s8) (v3),
        label={[font=\bfseries\small]above: Interactive Improvement (Steps 4--7)}] {};
\end{scope}

\end{tikzpicture}%
}
\caption{Overview of the LegalCheck pipeline, combining RAG with multi-stage CAG.}
\label{fig:legalcheck-workflow}
\Description{Overview of the LegalCheck pipeline, combining RAG with multi-stage CAG. The system first generates an initial draft advice letter (v1) by retrieving relevant items from a knowledge base and conditioning the LLM on the jurist’s case input, optionally after de-identifying personal information. The draft is then reviewed by the jurist, whose targeted feedback is used to construct a refinement prompt and produce a revised draft (v2), optionally with additional iterations and refreshed retrieval. After approval, personal identifiers can optionally be re-inserted before issuing the final letter.}
\end{figure*}

LegalCheck’s architecture follows a retrieval-augmented generation paradigm (RAG), extended with a multi-stage context-Augmented generation pipeline (CAG). Figure~\ref{fig:legalcheck-workflow} illustrates the workflow. When a new objection case is submitted, the system ingests the substantive case documentation, including, at minimum, the citizen’s objection letter and the enforcement officer’s report of the violation. In addition, the jurist is required to specify the intended dictum, that is, whether the objection should be upheld or rejected. Optionally, the jurist can add free-text steering advice that encodes case-specific priorities or nuances that are not captured elsewhere. These inputs form the core context for generation and also drive the retrieval query.

From this combined representation, the system constructs a textual query that summarizes the key facts and legal issues of the case. The query is encoded using a pre-trained embedding model (OpenAI’s text-embedding-ada-002, 1536-dimensional) to enable semantic search. The encoded query is matched against an internal knowledge base of previous cases and legal texts. In our implementation, this knowledge base comprises curated sets of historical legal advice letters from the domains included in the pilot, namely waste-fine objections and towing of vehicles, bicycles, and boats, together with the statutes and regulations most frequently cited in those letters.

Each document in the knowledge base is pre-processed into semantically meaningful chunks (typically by section) and stored with its embedding. We perform this semantic indexing offline; given the moderate size of the domain-specific datasets (ranging from approximately 150 to 14,000 cases per domain), vectors can be retrieved in-memory without a specialized vector database. The retrieval step returns the "Top-K" most relevant text chunks from previous cases, with "K" [50 - 200]. These sections typically include previous “Explanation” sections from similar objection letters, as well as snippets of legal provisions, ensuring that similar content of the new case is available to ground the AI’s drafting.

In summary, LegalCheck’s architecture consists of a robust knowledge retrieval layer feeding a GenAI, combined with a multi-stage CAG pipeline and explicit expert-in-the-loop control. The system combines the content of the new case, the collective memory of the department (past cases), and real-time expert judgment. In doing so, it produces advice letters that match or even exceed human-written quality, while substantially reducing the drafting effort. Section 4 describes how we evaluated these claims through a pilot study using both quantitative and qualitative metrics.

\subsection{Initial RAG+CAG Generation}

In the first generation stage, an LLM composes a draft advice letter using both the retrieved texts and the new case context. We use GPT-4o, deployed via the Azure OpenAI Service, as the primary generative engine \cite{openai2024hellogpt4o}. GPT-4o was chosen for its advanced capabilities in contextual understanding, fluent formal writing, and an extended context window (up to 128k tokens), which is beneficial when many retrieved documents are included. The system constructs a prompt that includes (a) the officer’s observation report, (b) the citizen’s objection letter, (c) any optional steering advice from the jurist, (d) a list of retrieved reference chunks, and (e) a predefined instruction template that guides the format and style of the output. The prompt template explicitly instructs the model to produce a structured legal advice letter that follows the municipality’s official style and uses the reference chunks provided to base its reasoning. For example, it is instructed to write the section “Explanation” in a formal, explanatory and empathetic tone that mirrors how exemplary human advisors write, to incorporate relevant laws and previous case details from the retrieved data, and to maintain phrasing conventions (such as how dates, case numbers, and fine amounts are referenced). By constraining the model in this way, we seek to ensure that the generated letter is not a free-form creation but a grounded, patterned output that adheres to customized legal writing norms. The model is run at low temperature (0.1) to reduce randomness, prioritizing factual consistency and stylistic uniformity over creative variance.

The outcome of this stage is a draft legal advice letter that typically includes an introduction, a section detailing the objection, a substantive “Explanation” section containing the legal assessment (this is the core where the model’s use of retrieved precedents comes in), and a conclusion stating the decision (uphold or reject the objection). Because the model is conditioned on actual passages from prior letters and legislation, it frequently produces explicit references or citations (e.g., to a specific article of a regulation or to the reasoning in similar earlier cases). This retrieval grounding substantially reduces hallucination and increases verifiability: AI key assertions can be traced back to real source documents provided in the prompt, rather than invented by the model. This design aligns with previous work on RAG~\cite{Lewis2020RAG} \cite{qi-etal-2024-model} \cite{nematov2025sourceattrib} and extends it to the domain of legal decision documents.

\subsection{Multi-Stage CAG Refinement}

The draft produced in the first generation stage is refined in an interactive CAG refinement stage. Through the interactive user interface, the jurist reads the AI-generated letter and can provide additional annotations, such as corrections, clarifications, or requests to add or emphasize certain arguments. When the jurist triggers the refinement function, the system formulates a new prompt to the LLM that includes the original AI draft, the underlying case documentation, the jurist’s annotations, and retrieved similar past cases to guide the refinement. The refine prompt instructs the model to revise the draft accordingly – e.g., insert the missing legal argument, adjust emphasis or tone, correct factual errors – while preserving the correct parts of the draft. We use the same GPT model for this refinement, with a custom prompt. We reuse the same model to ensure consistency in legal reasoning and writing style between v1 and v2, and because the refinement step is essentially a constrained rewrite of the same drafting task. In practice, a zero or one refinement pass is usually sufficient; multiple iterations are possible but rarely necessary.

This multi-stage CAG loop can be thought of as a form of \textit{iterative prompting}, where the model’s own output plus a new context (human feedback) serves as input for a second-generation pass. The process is analogous to a lawyer writing a memo and another lawyer marking it up with revision notes: LegalCheck then takes these notes and produces a revised draft. By integrating CAG, LegalCheck ensures that any case-specific nuances or expert intuitions that were not captured via retrieval alone can be injected into the final text. This proved valuable because even with high-quality retrieval, an AI draft might miss subtle contextual elements that a human deems important. The refinement step consistently produced improved drafts that satisfied the experts’ requirements. LegalCheck always keeps a human in the loop and never sends a letter directly to citizens or decision-makers. The jurist remains the final author, reviewing and approving the AI-assisted draft before it is issued. During early deployment, a second reviewer also double-checked AI-assisted letters as an additional safeguard.

\subsection{System Implementation}
The LegalCheck tool is implemented as a web application to integrate into the department’s workflow. A Flask-based Python web interface allows users to select a type of case and input the necessary case details: the objection text, the officer’s report, any additional documents (e.g., evidence attachments) and optional steering advice. Upon submission, the back-end loads the appropriate domain configuration. The system is modular, with separate pipelines for different case types. Each pipeline uses its own knowledge base and its own set of domain-specific contextual prompts. For each domain, the system configuration specifies which embedding index to use, which Azure OpenAI deployments to call for generation, refinement and embedding, and which document template to apply when exporting the final letter. Extending LegalCheck to a new legal domain (for example, building permit objections) therefore requires preparing a dataset of past cases, embedding them, and adding a corresponding configuration entry; the core logic remains unchanged. The retrieval is performed by computing cosine similarity between embeddings, and given the moderate dataset size, we did not require a dedicated vector search library because a custom Python implementation achieved millisecond-level latency even at the full scale of ~14k cases. We found this approach scalable, as no significant slow-down occurred when moving from a few hundred cases in prototyping to the full datasets. LLM calls (both initial draft and refine) are made via the Azure AI Service, which allows the deployment of GPT-4o in a secure cloud environment in compliance with the city’s IT policies.

\section{Methodology}

\subsection{Data Source and Knowledge Base}

Our data originate from historical legal objection cases handled by the municipality in the enforcement domains introduced in Section~1. From the central case library, we extracted cases from 2022 to 2025 that resulted in a formal written advice letter from the legal department. A typical case file contains the citizen’s objection letter, the official report by an enforcement officer (describing the alleged violation, for example dumping garbage at a specific time and location) and the written advice of the legal department to uphold or revoke the fine decision.

For retrieval, the advice letters served as the core corpus, since they contain the legal reasoning patterns and references that LegalCheck is designed to emulate. Across domains, we curated approximately 23,500 advice letters, with per-domain dataset sizes ranging from about 150 to 14,000 cases. Each letter was segmented into logical sections (e.g., \textit{Case Details}, \textit{Objection}, \textit{Hearing}, \textit{Explanation}, and \textit{Conclusion}). We applied section-based chunking so that the retrieval component can, for example, return only the “Explanation” segment of a previous letter, which is typically the most informative part to draft a new explanation. After chunking, we computed embeddings for each segment using text-embedding-ada-002. The resulting knowledge base is therefore a set of vector representations with pointers to their source text and metadata (such as case identifier and section type).

To comply with privacy and data-protection requirements, all previous case documents were anonymized or generalized before being used in prompts. Personal identifiers such as citizen names and addresses are replaced with placeholders, and any residual sensitive details are carefully filtered out in accordance with GDPR-compliant guidelines. All data used in the prompts are handled carefully to minimize the inclusion of personal identifiers and other sensitive information.

\subsection{Retrieval and Generation Process}
To generate a draft letter for a new case, the system first forms the query embedding from the concatenated objection and the officer’s report (and optionally a summary of any hearing transcript). Early experiments showed that summarizing the objection before embedding improved retrieval of similar cases, as it extracts the core issue. We adopted this as a preprocessing step: a separate LLM call produces a brief summary of the objection, which we then embed (this helps to abstract away verbose narratives while keeping key legal points). We first retrieve 50–200 similar objection sections using cosine similarity and then retrieve the corresponding explanation sections from the same cases, capturing how the jurists responded to those objections. This strategy directly yields material that can inform the new response. These retrieved chunks are concatenated into a structured prompt, after which the LLM, which has strong multilingual capabilities and handles Dutch legal text generation well, is then invoked to generate a first draft of the advice letter.

The prompt provides the case facts and the retrieved examples and instructs the model to write the “Explanation” section as a legally grounded analysis that mirrors the style and structure of the examples, addresses each argument and cites relevant regulations or past decisions. The prompt template and a concise system message that frames the model as a legal expert were iteratively refined to align the output with the department’s writing conventions and to steer it toward drafts comparable in quality, tone, and structure that are citizen-friendly to those of experienced jurists, with the final responsibility left to the human reviewer.

\subsection{Expert-in-the-Loop and CAG Refinement}
After the AI generates the draft letter (typically within 20–25 seconds, depending on the length of the context), the result is presented to the legal advisor via the interface. At this point, the advisor performs a qualitative review, verifying factual alignment, legal accuracy, and the clarity, tone, structure, and overall organization of the letter. In our evaluation phase, we log the time and nature of edits the jurists made. We also built the refinement (CAG) feature to help with this editing: the interface provides a text box for the jurist to input any instruction or correction and then press “Refine Draft”. 

\begin{figure}[!htbp]
\includegraphics[width=\columnwidth]{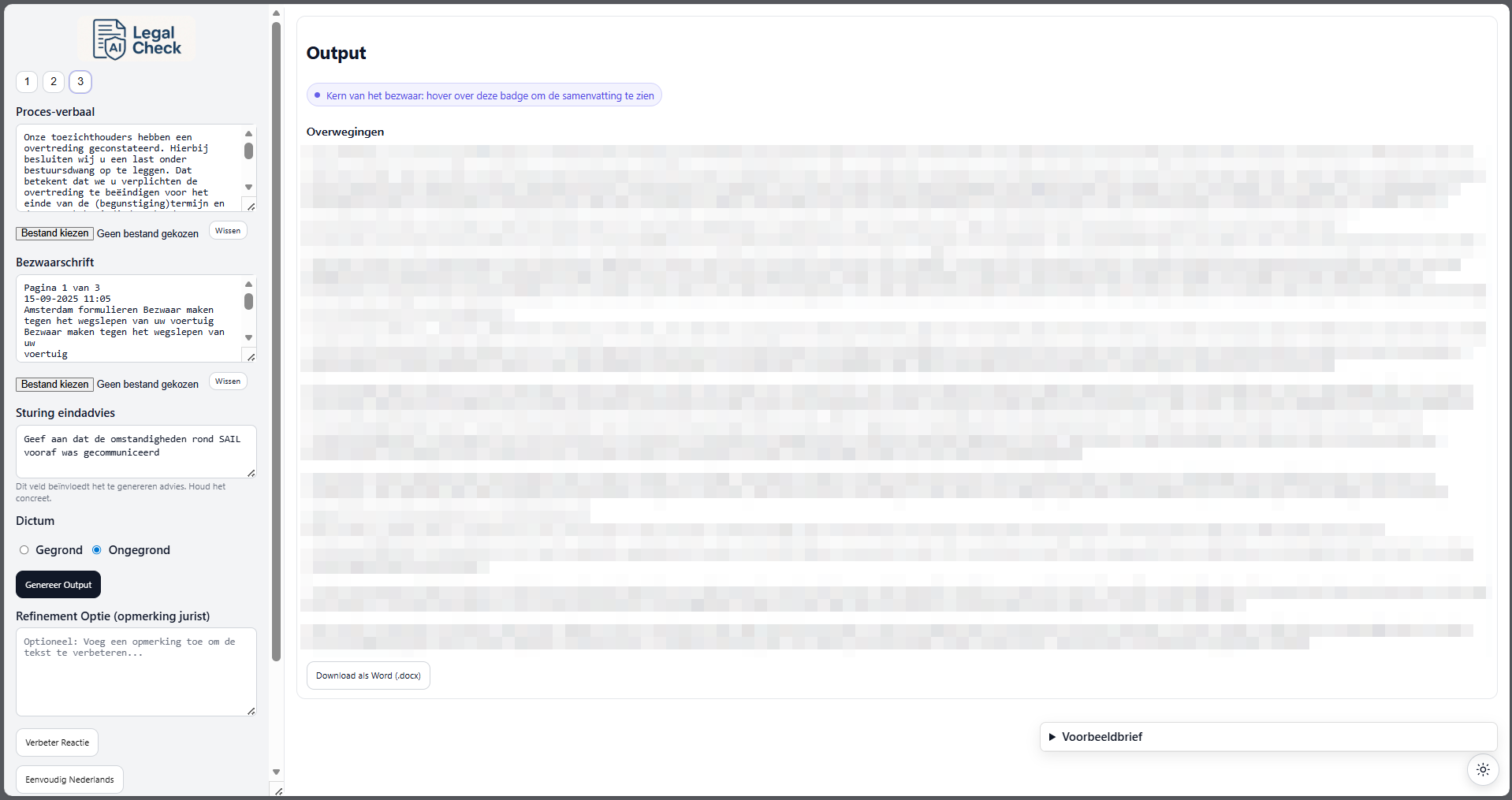}
  \caption{Screenshot of the LegalCheck user interface during drafting.}
  \label{fig:legalcheck-ui}
  \Description{Screenshot of the LegalCheck user interface during drafting. Most of the space is dedicated to the work in progress: elements of the case, draft of the letter. Input boxes allow the user to type its own text, for giving indications to the AI.}
\end{figure}

The system takes the original AI draft, the jurist’s remarks, and feeds them to GPT-4o in a special refine prompt (along with re-injection of the top reference chunks if needed for context). We found that this two-step generation significantly improved efficiency: jurists could simply highlight issues and let the AI fix them, rather than manually rewriting paragraphs. To evaluate CAG’s effectiveness, we tracked when jurists used the refinement and later asked their impressions. In all cases where it was used, the refinement feature produced a satisfactory revised draft, which the jurist then only needed to lightly proofread. It effectively handled additions of missing arguments or rephrasing according to the jurist’s guidance. The final reviewed (and refined, if applicable) letter is then approved by the jurist. A DOCX or PDF version of the letter can be downloaded, and a copy is logged together with metadata of which sources were used and what changes were made. This logging is critical for traceability and quality control, as every AI-generated document is stored with references to the information it relied on and how it was post-edited. This enables audits or explanations of AI suggestions if needed, providing a form of explainability through transparency: although the model’s internal reasoning is opaque, its outputs are traceable to their sources \cite{arrieta2020xai}.

\subsection{Evaluation Design}
We designed a mixed-method evaluation to assess (a) the quality and precision of LegalCheck’s output, (b) the efficiency gains (time saved), and (c) the user acceptance and trust in the system. The evaluation was conducted over a six-month period during the implementation of LegalCheck in the legal department’s workflow in the pilot domains.

\smallskip
\noindent\textbf{Quantitative Metrics:} We used several quantitative measures to compare AI-assisted drafting with traditional drafting:

\begin{itemize}[leftmargin=1.5em, itemsep=1pt, topsep=2pt, partopsep=2pt]
    \item \textbf{Similarity to Human-Written Letters:} For a sample of historical cases, we had LegalCheck generate a draft and compare them to the original human-written letters for the same cases. We measured an overlap ratio, defined as the percentage of AI-generated sentences or tokens that were retained in the final version after editing by jurists. On average, 81\% of AI-generated content was kept verbatim in the final letter, while jurists added only about 8\% new content. This suggests that the AI drafts were largely near-final and required only limited editing. To further quantify editing effort, we tracked the number of corrections made and found that, in many cases, jurists mainly fixed small errors (e.g., a reference detail) or added a few clarifying sentences.
    \item \textbf{Time Savings:} We asked approximately 20 participating jurists to log the time spent on each letter, broken down into (1) time to gather inputs and invoke AI, (2) time for AI generation, and (3) time to review and edit the AI-generated draft. We compared this with historical estimates of the time spent conducting legal research and drafting the letters from scratch, which averaged approximately 3 (waste), 6 (bicycles and vehicles) or 16 (boats) hours per case depending on the domain. The results showed an overall reduction of roughly 50–70\% in total time for straightforward cases. In absolute terms, work that previously required  ~6 hours could be completed in ~2–3 hours with the help of LegalCheck. This aligns with anecdotal reports from jurists that having a solid draft “instantly” enabled them to complete letters much more quickly, particularly for recurring objection patterns. We further extrapolated the potential impact across the pilot domains: given ~5,800 objections per year, annual time savings are around 12k–17k hours of jurist work, corresponding to roughly 10–14 full-time legal staff in workload. At an average loaded cost of €100 per hour for a legal officer, this equates to an estimated €1.2–1.7 million in notional annual cost savings for the department. Although these figures assume full deployment and that all saved time is effectively realized as additional capacity, they underscore the substantial business impact of the system. Instead of doing legal research from scratch, jurists can reallocate time to more complex cases or direct interactions with citizens.
    \item \textbf{Content Accuracy and Consistency:} We evaluated whether AI letters correctly cited relevant laws and applied appropriate legal reasoning. In our sample of approximately 250 cases, the AI draft omitted a context-specific detail in a small number of instances, which the jurist then corrected during the refinement step or in the final edit.\\ We measured the retention of key facts by checking whether each AI draft contained the critical factual elements of the case (such as date of offense, type of violation, and location). In > 99\% of the drafts, all key facts were present. The consistency of legal reasoning (e.g., following the standard structure of proportionality analysis, citing relevant municipal ordinances) was maintained across drafts, which helped to standardize letters. A notable side effect was that AI-drafted letters tended to be more comprehensive than human-written ones. In 68\% of the cases, AI letters were longer, with on average about 41\% more content, often in the form of additional justifications and responses to more arguments raised in the objection. In 32\% of cases, human letters were longer (by about 12\%). This suggests that the model, drawing on a broad range of previous cases, sometimes introduced arguments that an individual jurist might not have recalled. Jurists generally viewed this as positive, as one noted, \textit{“the AI output sometimes raised a point I had not initially considered, which made the letter stronger.”}
\end{itemize}

\smallskip
\noindent\textbf{Qualitative Evaluation}: We gathered feedback through surveys and interviews with the legal officers who piloted the tool, as well as additional stakeholders (other jurists, department managers, and subject-matter experts in legal technology). The feedback was positive with respect to time and quality. Jurists reported that LegalCheck’s draft letters required only minimal editing and that the tool \textit{“clearly speeds up the handling of cases”}, especially for common objection scenarios. They noted that AI was adept at maintaining formal tone and even injecting empathy appropriately for example acknowledging the citizen’s perspective politely. Some were initially skeptical about trusting an AI for legal writing, but after reviewing the output, their trust increased. A jurist said \textit{“I was looking for mistakes, but I mostly found correct references and consistent reasoning after a few letters I became more confident in it.”} Indeed, a couple of jurists who started as cautious evaluators became enthusiastic adopters at the end of the pilot, actively integrating the tool into their daily workflow. This echoes the general observations that practitioners are more willing to trust AI as a support tool once they see that it works reliably \cite{Venkatesh2003Acceptance}.

We explicitly solicited criticisms and concerns. The primary concern was the need for oversight: jurists were clear that AI, no matter how good it is, should not be used without human review (a stance we fully endorse and built into the process). They felt comfortable with the tool as \textit{“the first drafter”}, but they asserted that final judgments must remain with human jurists, especially for borderline cases or ones involving discretion. This aligns with our expert-in-the-loop design and with public sector AI guidelines that stress human final decision-making for legal matters. Another point raised was about explainability: if challenged, could the department explain how the letter was generated? We addressed this by logging the sources and the prompt for each case, effectively creating an audit trail of the AI’s inputs and outputs. This way, if needed, an expert can reconstruct what information LegalCheck used and why certain language or citations appear, providing some level of explanation despite the LLM being a \textit{“black box.”} Jurists also suggested improvements such as integrating image analysis (some cases include photo evidence, which is currently analyzed with AI in early experiments) and expanding to other types of cases. We have since started extending LegalCheck to the other domains mentioned, with similarly promising early results.

We summarize the key evaluation outcomes in Table 1, which includes metrics such as content retention, time saved per document, and user satisfaction ratings. In general, the evaluation confirmed that LegalCheck meets its goals: it significantly accelerates the drafting process while maintaining high legal quality, and it has been well-received by the legal staff as a trustworthy assistant rather than a replacement \cite{haesevoets2025aiadoption}. In the next section, we discuss in more depth the implications of these findings for legal reasoning, the role of the lawyer, and responsible AI use in law.

\begin{table}[t]
\label{tab:evaluation}
\centering
\small
\begin{tabular}{p{0.32\columnwidth} p{0.40\columnwidth} p{0.20\columnwidth}}
\hline
\textbf{Evaluation aspect} & \textbf{Metric (definition)} & \textbf{Result} \\
\hline
Content coverage &
Recall &
93\% \\
Content relevance &
Precision &
91\% \\
Overall quality &
F1-score &
92\% \\
Draft usability &
Retention ratio (share of AI draft kept verbatim after jurist editing) &
81\% \\
Editing efficiency &
Time-saved ratio for drafting stage &
92\% \\
End-to-end efficiency &
Reduction in total case handling time &
50--70\%\ \\
\hline
\end{tabular}
\caption{Evaluation results of LegalCheck compared to human-written advice letters.}
\vspace{-7mm}
\end{table}

\section{Discussion}
\subsection{Impact on Legal Reasoning and Professional Roles}

\smallskip
\noindent\textbf{Effects on Legal Reasoning and Consistency:} The deployment of LegalCheck offers insight into how AI can augment legal reasoning in practice. By design, the system employs a case-based reasoning approach: it draws on past cases to inform the solution of a new case. This aligns with how lawyers traditionally work, by finding precedents and analogies, but automates both the retrieval and the initial application of those precedents. An immediate impact is increased consistency in legal advice letters. Since AI’s output in each case is based on collective knowledge of previous decisions, letters tend to reflect a more uniform application of rules and arguments throughout the department. This can reduce inadvertent disparities between the way different staff write advice on similar cases. It effectively standardizes best practices, as AI tends to include the strongest and most citizen-oriented arguments and citations present in the knowledge base. For legal professionals, this means that their role shifts slightly from being sole content generators to serving as content curators and reviewers \cite{hashem2025mappingpotential}. They spend less time on the assembly of legal arguments and more on reviewing the AI’s draft, checking nuances, and making judgment calls on edge cases. Importantly, the jurists remain the ultimate decision-makers: the AI does not alter the outcome of a case (it does not decide to uphold or overturn a fine on its own; it only drafts the rationale for the decision that the human inputs as the intended outcome). 

\smallskip
\noindent\textbf{Effects on Professional Roles, Skills and Learning:} In our use case, the legal outcome (uphold/waive the fine) was determined by the human based on the facts; LegalCheck then assists in formulating the justification for that outcome. This is a crucial distinction which maintains accountability with the human official. From the interviews, lawyers felt empowered rather than threatened by the tool, as it saved them from drudgery (“no more starting from a blank page”) and from manually stitching together boilerplate text blocks. This allowed them to focus on higher-level analysis and on difficult cases that truly require their expertise.

This echoes themes in the legal AI literature that envision AI systems handling a substantial portion of routine drafting and research, allowing lawyers to focus on more complex high-stakes tasks \cite{Susskind2019Tomorrow}. Jurists already use fixed text blocks and templates in their daily work, but these boilerplates never fully capture the specific constellation of facts and arguments in an individual case and must then be manually adapted. LegalCheck goes beyond such template-based support by selecting and rewriting relevant building blocks in light of the full case context, producing a coherent, case-specific explanation. We also observed that junior staff benefitted from AI drafts as a learning aid; a junior jurist said it was like having a \textit{“reference draft”} from a seasoned colleague, which helped them improve their own writing by example. Over time, as such tools become common, the skill set for legal professionals may evolve, and proficiency in working with AI (prompting it effectively and critically evaluating its output) will become important, much like proficiency with legal research databases is today.

\smallskip
\noindent\textbf{Trust, Adoption and External Perception:} Another significant impact is on the efficiency of legal research. Tasks that previously required combing through case files are now completed in seconds by the retrieval module. In the pilot, jurists noted that they discovered relevant past cases that they would not have found or remembered themselves because the semantic search could surface matches even when terminology differed. This improved the thoroughness of legal reasoning, as the advice letters could reference more precedents. Indeed, LegalCheck integrates a legal research engine and a drafting assistant into one seamless workflow.

Trust in AI-supported decision-making is paramount in the legal domain. At the outset, we encountered healthy skepticism from some jurists, who were concerned that AI might make subtle errors or produce text that they could not fully trust \cite{alon-barkat2023humanai}. Through the expert-in-the-loop design and the consistent quality of the output, this skepticism was largely overcome. Key trust-building features were the inclusion of sources (which made jurists comfortable that they could verify any statement) and the requirement that they approve every letter (maintaining a sense of control). This confirms research suggesting that users prefer AI as a support tool rather than an autonomous agent in professional settings. Our approach led jurists to treat AI almost like a colleague or first-draft writer: useful but always subject to review. During the pilot, we saw a progression: initial caution gave way to routine use, and by the end some jurists remarked that they \textit{“would not want to go back”} to the old way of drafting. This level of acceptance is encouraging, but was facilitated by transparent communication about the capabilities and limitations of AI, training sessions, and management support. Department leaders emphasized that AI was introduced to support lawyers rather than to evaluate them, and this cultural framework helped reduce fears about automation. We recommend any similar AI introduction follow a participatory approach: involve end-users early, allow them to test and give feedback, and iterate the tool accordingly.

One question often raised is how the use of AI in legal drafting might be perceived by external parties (citizens or judges). In our context, the final letters did not explicitly disclose that AI was involved, since letters are always reviewed by a human. The content was virtually indistinguishable from human-written letters. Thus, there was no observable change in the way letters would appear in subsequent legal proceedings. In the long run, should AI drafting become commonplace, there may be discussions about whether to disclose AI assistance for transparency.

\subsection{Responsible AI and Compliance Considerations}

\smallskip
\noindent\textbf{Privacy and Data Protection:} Implementing LegalCheck in a public sector context required careful attention to ethical and legal compliance principles from the beginning. We address privacy by ensuring that sensitive personal data are protected beyond what is necessary \cite{eu2016gdpr}. Where feasible, data sent to the OpenAI API is minimized and stripped of personal identifiers. In addition, under the agreement of the municipality with the cloud provider, API data is not used to train models and are processed under the configured Azure OpenAI privacy and compliance settings, including encryption, access controls, and logging/retention controls. Additionally, because the knowledge base consists of internal documents, we hosted the embedding and retrieval components on a secure municipal server. For each request, only the specific retrieved passages needed for answering are sent to Azure for LLM inference, rather than the full knowledge base. 

\smallskip
\noindent\textbf{Transparency, Accountability and Risk Management:} On the ethical and transparency front, the traceability of AI decisions is a key focus. As described, the system logs sources and outputs. This not only aids internal audits, but also contributes to explainability: even if we cannot explain the neural mechanisms of the GPT model, we can explain what information it used and what instructions it received, which is often sufficient in a legal setting to justify the basis of an advice. To further enhance transparency, we structured the AI’s output to reflect a logical legal argument, making it easier for a human reviewer to follow and for an external observer to understand the reasoning. In essence, the role of AI is disclosed internally and all drafts are reviewed, providing accountability \cite{ai4citizens2025ethicalleaflet}.

We also proactively aligned the project with the requirements of the forthcoming EU AI Act, although the law is not yet in force in 2025 \cite{eu2024aiact}. We performed a LegalCheck risk classification and determined that it falls into a likely “low-risk” category: an AI system for administrative decision support (not fully autonomous decision-making, not biometrics or surveillance, etc.) \cite{sheehy_ng_2024_challenges_ai}. Low-risk does not mean no risk, so we implemented voluntary safeguards: documentation of the purpose and limitations, keeping jurists in charge of decisions, and monitoring for errors or biases. The department’s compliance team reviewed LegalCheck and approved the use of the OpenAI model through Azure, noting that it should use minimal personal data and that legal validity checks must remain in place. We also considered bias/fairness: since the system relies on past decisions, there is a risk that it could perpetuate any biases in those decisions. To mitigate this, we had legal experts review a sample of AI-generated letters for signs of inconsistent treatment. No issues were found in our pilot domains, but this is something to monitor in other legal domains (e.g., if applied to social benefits cases, one would watch for any bias against certain groups). Our governance plan includes periodic audits of outputs if the system is scaled up.

\smallskip
\noindent\textbf{User Training and Organizational Measures:} Finally, we emphasize user training and the organizational measures taken. All jurists using LegalCheck attended training sessions on how the system works, its intended use, and how to interpret its suggestions. We created a user guide that explicitly instructs: \textit{“AI is a drafting aid. You must read and correct the AI draft as you would review a junior colleague draft. You are responsible for the final content.”} By setting clear expectations, we ensured that the tool is used properly and does not erode the sense of responsibility. We also set up a feedback channel so that users could report any problematic AI output to developers and data scientists. During the pilot, a few minor bugs related to formatting or missing context were identified and promptly fixed, demonstrating the importance of having a responsive support process.

In summary, LegalCheck’s implementation was accompanied by a comprehensive and responsible AI framework: privacy-by-design, human oversight, traceability, compliance with regulatory guidelines, and multidisciplinary governance oversight \cite{vatamanu2025integrating}. We believe that this is a model for how to introduce AI into government workflows in a manner that maintains public trust \cite{weerts2025generativeaipublicadmin}. This included leadership support and strong cross-functional collaboration across legal, data, ethics, and IT teams\cite{kotter1996leadingchange}.

\section{Conclusion}

\smallskip
\noindent\textbf{Academic and Practical Implications:} We presented LegalCheck, a retrieval-augmented generation system with a multi-stage context-augmented generation pipeline to support municipal jurists in drafting objection advice letters. In a pilot in the City of Amsterdam, LegalCheck produced legally grounded drafts substantially faster than manual drafting while keeping the jurist fully responsible for the final content. \cite{sengupta2025legalsectoradaptability}.

For the AI-and-law research community, LegalCheck contributes a case study of deploying a RAG-based system, enriched with multi-stage CAG, in an environment that demands high accountability. The results underscore the importance of domain-specific and organization-specific data: instead of relying on a generic GPT deployment, our model was “educated” in the municipality’s own jurisprudence, which was crucial to its performance. Conceptually, we contribute a multi-stage form of context-augmented generation for legal text: the model conditions on rich case context at the initial generation stage and then uses expert feedback and patterns from analogous past cases in subsequent refinement passes. This opens avenues for further work on interactive, human-guided legal text generation, in which LLMs operate as adaptive drafting partners rather than one-shot generators. Our approach resonates with current ideas of AI “reasoning models” coupled with law retrieval systems and provides a concrete implementation of such architectures in a public-sector setting. In addition, our evaluation design, which combines quantitative text-level metrics with qualitative insights on trust, workflow integration, and perceived value, offers a template for assessing legal AI tools beyond standard NLP benchmarks.

\smallskip
\noindent\textbf{Limitations:} Our study was conducted in a single municipal organization, in four specific enforcement domains, and with a relatively small group of legal professionals. Generalization to other legal domains (e.g., criminal law, civil litigation) therefore  requires caution. Different document types (briefs, judicial opinions, contracts) have their own conventions and ethical stakes. However, the architecture of LegalCheck is domain-agnostic; it can be reconfigured for other use cases if a relevant case repository is available. Another limitation is reliance on a third-party LLM. We mitigated these risks by using Azure’s enterprise-grade offering and abstracting the LLM layer in our code, making the system model-agnostic. Indeed, as open-source LLMs improve, a future iteration might use an in-house model to reduce dependency. We did not systematically examine the longer-term effects on employment, professional identity, or perceived fairness from the citizen’s perspective. Our focus was on augmenting a team facing staff shortages and growing backlogs, and the tool was perceived to reduce workload pressure rather than threaten jobs. However, as these tools become more widespread, an ongoing dialog with stakeholders, including potential citizens, will be needed to navigate the balance between efficiency gains and (only) the human touch in public services.

\smallskip
\noindent\textbf{Future Work:} We are extending LegalCheck to additional case types (e.g. objections to permits for short-term vacation rentals and building permits) to test scalability. Pilots on current case types have shown similar results, although prompt configurations and retrieval filters required adjustment to reflect domain-specific terminology and decision patterns. A technical priority is to further develop image analysis using the multimodal GPT-4o capabilities of GPT-4o, using the vision-language model to produce structured descriptions of such evidence that can be incorporated into the textual prompts for both the RAG and CAG stages of the pipeline. We also intend to develop a more advanced user interface that supports inline editing (that the jurist can edit the draft directly and the system logs those edits to learn from them). 
On the research side, we aim to formally evaluate the multi-stage CAG process itself, by comparing one-pass RAG generation with RAG plus CAG refinement in a blind study, asking jurists to rate letters on legal quality, completeness, clarity, and citizen-friendliness without knowing which condition produced them. We also intend to monitor longer-term organizational effects, such as changes in throughput, backlog levels, perceived workload, and the evolution of roles and skills within the legal team. These sociotechnical aspects are essential to understand the broader impact of AI-assisted writing on public services and the transformation of legal work.

With regard to reproducibility and collaboration, the underlying case data are confidential, but we are exploring the creation of an anonymized synthetic dataset that mirrors the structure and logic of the real objection letters without disclosing sensitive content. This resource could serve as a reference corpus for the RAG and CAG methods in the generation of legal texts. We have modularized key components of the system, including the retrieval module, the prompt-construction logic, evaluation scripts,  and the configuration layer for domain-specific pipelines, and we are currently exploring how these artifacts can be shared for research purposes. By sharing these artifacts, we hope to facilitate further work on legal AI systems that are both technically rigorous and responsibly deployed.


\section*{Usage of Generative AI}
Generative AI tools were used to refine the authors writing of this paper and to generate figures or tables. 

\begin{acks}
Generative AI tools were used to refine the authors writing of this paper and to generate figures or tables. 
\end{acks}

\bibliographystyle{ACM-Reference-Format}
\bibliography{biblio}

@String{Computing = "Computing" }

@String{Computer = "{IEEE} Computer" }

@misc{VNG2024Pilot,
  author       = {Vereniging van Nederlandse Gemeenten (VNG)},
  title        = {{Pilot big data \& AI-tools voor effici\"entere afhandeling bezwaarschriften}},
  howpublished = {VNG website (Oct~29, 2024)},
  note         = {URL: \url{https://vng.nl/artikelen/pilot-big-data-ai-tools-voor-efficientere-afhandeling-bezwaarschriften}},
  year         = {2024}
}

@article{Davenport2018RealWorld,
  author  = {Davenport, Thomas H. and Ronanki, Rajeev},
  title   = {Artificial Intelligence for the Real World},
  journal = {Harvard Business Review},
  volume  = {96},
  number  = {1},
  pages   = {108--116},
  year    = {2018}
}

@article{Aletras2016ECtHR,
  author    = {Aletras, Nikolaos and Tsarapatsanis, Dimitrios and Preotiuc-Pietro, Daniel and Lampos, Vasileios},
  title     = {Predicting Judicial Decisions of the European Court of Human Rights: A Natural Language Processing Perspective},
  journal   = {PeerJ Computer Science},
  volume    = {2},
  pages     = {e93},
  year      = {2016},
  doi       = {10.7717/peerj-cs.93}
}

@article{Katz2017SCOTUS,
  author    = {Katz, Daniel Martin and Bommarito, Michael J. and Blackman, Josh},
  title     = {A General Approach for Predicting the Behavior of the Supreme Court of the United States},
  journal   = {PLoS ONE},
  volume    = {12},
  number    = {4},
  pages     = {e0174698},
  year      = {2017},
  doi       = {10.1371/journal.pone.0174698}
}

@techreport{Schwarcz2025AIPowered,
  author      = {Schwarcz, Daniel and Manning, Sam and Barry, Patrick J. and Cleveland, David R. and Prescott, J. J. and Rich, Beverly},
  title       = {{AI-Powered Lawyering: AI Reasoning Models, Retrieval Augmented Generation, and the Future of Legal Practice}},
  institution = {Minnesota Legal Studies Research Paper No.~25-16 (SSRN)},
  year        = {2025},
  note        = {Available at SSRN: \url{https://ssrn.com/abstract=5162111}}
}

@inproceedings{Lewis2020RAG,
  author    = {Lewis, Patrick and Perez, Ethan and Piktus, Aleksandra and Petroni, Fabio and Karpukhin, Vladimir and Goyal, Naman and K\"uttler, Heinrich and Lewis, Mike and Yih, Wen-tau and Rockt\"aschel, Tim and Riedel, Sebastian and Kiela, Douwe},
  title     = {{Retrieval-Augmented Generation for Knowledge-Intensive NLP Tasks}},
  booktitle = {Advances in Neural Information Processing Systems 33},
  pages     = {9459--9474},
  year      = {2020}
}

@article{Venkatesh2003Acceptance,
  author  = {Venkatesh, Viswanath and Morris, Michael G. and Davis, Gordon B. and Davis, Fred D.},
  title   = {User Acceptance of Information Technology: Toward a Unified View},
  journal = {MIS Quarterly},
  volume  = {27},
  number  = {3},
  pages   = {425--478},
  year    = {2003}
}

@book{Susskind2019Tomorrow,
  author    = {Susskind, Richard},
  title     = {{Tomorrow's Lawyers: An Introduction to Your Future}},
  edition   = {2nd},
  publisher = {Oxford University Press},
  year      = {2019}
}

@article{eu2024aiact,
  title={Regulation (EU) 2024/1689 laying down harmonised rules on artificial intelligence (AI Act)},
  author={European Union},
  journal={Official Journal of the European Union},
  year={2024}
}

@misc{amsterdam2024visionai,
  author       = {{Gemeente Amsterdam}},
  title        = {Amsterdam’s vision on AI (English version)},
  year         = {2024},
  url          = {https://www.amsterdam.nl/innovatie/amsterdamse-visie-ai/}
}

@misc{chienkim2024legalaid,
  author       = {Chien, Colleen V. and Kim, M},
  title        = {Generative AI and Legal Aid: Results from a Field Study and 100 Use Cases to Bridge the Access to Justice Gap},
  year         = {2024},
  month        = mar,
  day          = {14},
  howpublished = {SSRN Working Paper (UC Berkeley Public Law Research Paper; forthcoming in Loyola of Los Angeles Law Review)},
  url          = {https://ssrn.com/abstract=4733061}
}

@mastersthesis{wrzesniowska2023canaimakeacase,
  author       = {Wrzesniowska, L.},
  title        = {Can AI make a case? AI vs. lawyer in the Dutch legal context},
  school       = {University of Amsterdam},
  year         = {2023},
  note         = {Master's thesis; later appearing in The International Journal of Law, Ethics, and Technology}
}

@misc{europeancommission2019trustworthyai,
  author       = {{European Commission: Directorate-General for Communications Networks, Content and Technology} and {High-Level Expert Group on Artificial Intelligence}},
  title        = {Ethics Guidelines for Trustworthy AI},
  year         = {2019},
  publisher    = {Publications Office of the European Union},
  doi          = {10.2759/346720},
  url          = {https://data.europa.eu/doi/10.2759/346720}
}

@article{surden2019aioverview,
  author  = {Surden, Harry},
  title   = {Artificial Intelligence and Law: An Overview},
  journal = {Georgia State University Law Review},
  year    = {2019},
  volume  = {35},
  number  = {4},
  pages   = {1305--1337}
}

@book{ashley2017ailegalanalytics,
  author    = {Ashley, Kevin D.},
  title     = {Artificial Intelligence and Legal Analytics: New Tools for Law Practice in the Digital Age},
  publisher = {Cambridge University Press},
  year      = {2017}
}

@inproceedings{lincheng2024legaldrafting,
  author    = {Lin, Chih-Hao and Cheng, Pei-Ju},
  title     = {Legal documents drafting with fine-tuned pre-trained large language model},
  booktitle = {Proceedings of the 12th International Conference on Software Engineering \& Trends (SE 2024)},
  year      = {2024},
  address   = {Copenhagen, Denmark},
  doi       = {10.48550/arXiv.2406.04202},
}

@misc{wiratunga2024cbrrag,
  author       = {Wiratunga, N. and Abeyratne, R. and Jayawardena, L. and Martin, K. and Massie, S. and Nkisi-Orji, I. and Weerasinghe, R. and Liret, A. and Fleisch, B.},
  title        = {CBR-RAG: Case-Based reasoning for retrieval augmented generation in LLMs for legal question answering},
  year         = {2024},
  howpublished = {arXiv preprint},
  doi          = {10.48550/arXiv.2404.04302},
  url          = {https://arxiv.org/abs/2404.04302}
}

@inproceedings{amershi2019guidelines,
  author    = {Amershi, Saleema and Weld, Daniel and Vorvoreanu, Mihaela and Fourney, Adam and Nushi, Besmira and Collisson, Penny and Suh, Jina and Iqbal, Shamsi and Bennett, Paul N. and Inkpen, Kori and Teevan, Jaime},
  title     = {Guidelines for Human--AI Interaction},
  booktitle = {Proceedings of the 2019 CHI Conference on Human Factors in Computing Systems (CHI '19)},
  year      = {2019},
  publisher = {ACM},
  doi       = {10.1145/3290605.3300233},
  note      = {Article 3, pp. 1--13}
}

@misc{magesh2024hallucinationfree,
  author       = {Magesh, Varun and Surani, F. and Dahl, M. and Suzgun, Mirac and Manning, Christopher D. and Ho, Daniel E.},
  title        = {Hallucination-free? Assessing the reliability of leading AI legal research tools},
  year         = {2024},
  howpublished = {arXiv preprint},
  doi          = {10.48550/arXiv.2405.20362},
  url          = {https://arxiv.org/abs/2405.20362}
}

@inproceedings{holstein2019improvingfairness,
  author    = {Holstein, Kenneth and Wortman Vaughan, Jennifer and Daum{\'e} III, Hal and Dudik, Miro and Wallach, Hanna},
  title     = {Improving Fairness in Machine Learning Systems: What Do Industry Practitioners Need?},
  booktitle = {Proceedings of the 2019 CHI Conference on Human Factors in Computing Systems (CHI '19)},
  year      = {2019},
  publisher = {ACM},
  doi       = {10.1145/3290605.3300830},
  note      = {pp. 1--16}
}

@article{haesevoets2025aiadoption,
  author  = {Haesevoets, T. and Verschuere, B. and Roets, A.},
  title   = {AI adoption in public administration: Perspectives of public sector managers and public sector non-managerial employees},
  journal = {Government Information Quarterly},
  year    = {2025},
  volume  = {42},
  number  = {2},
  pages   = {102029},
  doi     = {10.1016/j.giq.2025.102029}
}

@misc{ai4citizens2025ethicalleaflet,
  author       = {{AI4Citizens}},
  title        = {Ethical leaflet: Get transparency about the moral implications of technology used},
  year         = {2025},
  month        = jun,
  day          = {19},
  howpublished = {Interreg Europe -- Good practices},
  url          = {https://www.interregeurope.eu/good-practices/ethical-leaflet-get-transparency-about-moral-implications-of-technology-used}
}

@article{arrieta2020xai,
  author  = {Arrieta, Alejandro Barredo and D{\'i}az-Rodr{\'i}guez, Natalia and Del Ser, Javier and Bennetot, Adrien and Tabik, Siham and Barbado, Alberto and Garc{\'i}a, Salvador and Gil-L{\'o}pez, Sergio and Molina, Daniel and Benjamins, Richard and Chatila, Raja and Herrera, Francisco},
  title   = {Explainable artificial intelligence (XAI): Concepts, taxonomies, opportunities and challenges toward responsible AI},
  journal = {Information Fusion},
  year    = {2020},
  volume  = {58},
  pages   = {82--115},
  doi     = {10.1016/j.inffus.2019.12.012}
}

@techreport{bommasani2021foundationmodels,
  author      = {Bommasani, Rishi and others},
  title       = {On the Opportunities and Risks of Foundation Models},
  institution = {Stanford Institute for Human-Centered Artificial Intelligence},
  year        = {2021},
  type        = {Technical Report},
  note        = {arXiv:2108.07258},
  url         = {https://arxiv.org/abs/2108.07258}
}

@misc{eu2016gdpr,
  author       = {{European Union}},
  title        = {Regulation (EU) 2016/679 (General Data Protection Regulation)},
  year         = {2016},
  howpublished = {Official Journal of the European Union, L 119, 1--88},
  url          = {https://eur-lex.europa.eu/eli/reg/2016/679/oj}
}

@article{faraj2018learningalgorithm,
  author  = {Faraj, Samer and Pachidi, Stella and Sayegh, Karim},
  title   = {Working and organizing in the age of the learning algorithm},
  journal = {Information and Organization},
  year    = {2018},
  volume  = {28},
  number  = {1},
  pages   = {62--70},
  doi     = {10.1016/j.infoandorg.2018.02.005}
}

@misc{hashem2025mappingpotential,
  author      = {Hashem, Yousra and Bright, Jonathan and Chakraborty, Shreya and Onslow, Kait and Francis, James and Poletav, A. and Esnaashari, S.},
  title       = {Mapping the Potential: Generative AI and Public Sector Work. Using time use data to identify opportunities for AI adoption in Great Britain’s public sector},
  institution = {The Alan Turing Institute},
  year        = {2025},
  url         = {https://www.turing.ac.uk/sites/default/files/2025-05/ons_tus_final_report.pdf}
}

@book{kotter1996leadingchange,
  author    = {Kotter, John P.},
  title     = {Leading Change},
  publisher = {Harvard Business School Press},
  year      = {1996}
}

@misc{openai2024hellogpt4o,
  author       = {{OpenAI}},
  title        = {Hello GPT-4o},
  year         = {2024},
  month        = may,
  day          = {13},
  url          = {https://openai.com/nl-NL/index/hello-gpt-4o/}
}

@misc{pwc2023dutchjobsgenai,
  author       = {{PwC}},
  title        = {Half of Dutch jobs might be significantly changed by generative AI},
  year         = {2023},
  month        = sep,
  day          = {28},
  howpublished = {PwC Netherlands},
  url          = {https://www.pwc.nl/en/insights-and-publications/themes/the-future-of-work/half-of-dutch-jobs-might-be-significantly-changed-by-generative-ai.html}
}

@misc{sengupta2025legalsectoradaptability,
  author       = {SenGupta, R.},
  title        = {Legal sector’s adaptability proves the key to success},
  year         = {2025},
  month        = jun,
  day          = {25},
  howpublished = {Financial Times},
  url          = {https://www.ft.com/content/6df512fe-c1a7-4ed1-be3d-d3dc5e5b2944}
}

@misc{thomsonreuters2025lessdrudge,
  author       = {{Thomson Reuters}},
  title        = {Less drudge, more expertise: How AI is redefining the future of legal professionals in Australia},
  year         = {2025},
  month        = jul,
  day          = {21},
  howpublished = {The Guardian (Thomson Reuters AI Futures)},
  url          = {https://www.theguardian.com/thomson-reuters-ai-futures/2025/jul/21/less-drudge-more-expertise-how-ai-is-redefining-the-future-of-legal-professionals-in-australia}
}

@article{vatamanu2025integrating,
  author  = {Vatamanu, Andrei F. and Tofan, M.},
  title   = {Integrating artificial intelligence into public administration: Challenges and vulnerabilities},
  journal = {Administrative Sciences},
  year    = {2025},
  volume  = {15},
  number  = {4},
  pages   = {149},
  doi     = {10.3390/admsci15040149}
}

@article{weerts2025generativeaipublicadmin,
  author  = {Weerts, S.},
  title   = {Generative AI in public administration in light of the regulatory awakening in the US and EU},
  journal = {Cambridge Forum on AI: Law and Governance},
  year    = {2025},
  pages   = {e3},
  doi     = {10.1017/cfl.2024.10}
}

@misc{shao2025when,
  title         = {When Large Language Models Meet Law: Dual-Lens Taxonomy, Technical Advances, and Ethical Governance},
  author        = {Shao, Peizhang and Xu, Linrui and Wang, Jinxi and Zhou, Wei and Wu, Xingyu},
  year          = {2025},
  month         = jul,
  doi           = {10.48550/arXiv.2507.07748},
  url           = {https://arxiv.org/abs/2507.07748}
}

@article{alon-barkat2023humanai,
  title   = {Human--AI Interactions in Public Sector Decision Making: {``Automation Bias''} and {``Selective Adherence''} to Algorithmic Advice},
  author  = {Alon-Barkat, Saar and Busuioc, Madalina},
  journal = {Journal of Public Administration Research and Theory},
  year    = {2023},
  volume  = {33},
  number  = {1},
  pages   = {153--169},
  month   = jan,
  doi     = {10.1093/jopart/muac007},
  url     = {https://doi.org/10.1093/jopart/muac007}
}

@article{sheehy_ng_2024_challenges_ai,
  title   = {The Challenges of AI-Decision-Making in Government and Administrative Law: A Proposal for Regulatory Design},
  author  = {Sheehy, Benedict and Ng, Yee-Fui},
  journal = {Indiana Law Review},
  year    = {2024},
  volume  = {57},
  number  = {3},
  pages   = {665--699},
  month   = jun,
  date    = {2024-06-10},
  doi     = {10.18060/28360},
  url     = {https://journals.indianapolis.iu.edu/index.php/inlawrev/article/view/28360}
}

@inproceedings{qi-etal-2024-model,
  author    = {Qi, Jirui and Sarti, Gabriele and Fern{\'a}ndez, Raquel and Bisazza, Arianna},
  title     = {Model Internals-based Answer Attribution for Trustworthy Retrieval-Augmented Generation},
  booktitle = {Proceedings of the 2024 Conference on Empirical Methods in Natural Language Processing},
  year      = {2024},
  address   = {Miami, Florida, USA},
  publisher = {Association for Computational Linguistics},
  url       = {https://aclanthology.org/2024.emnlp-main.347/},
  doi       = {10.18653/v1/2024.emnlp-main.347}
}

@article{nematov2025sourceattrib,
  author  = {Nematov, Ikhtiyor and Kalai, Tarik and Kuzmenko, Elizaveta and Fugagnoli, Gabriele and Sacharidis, Dimitris and Hose, Katja and Sagi, Tomer},
  title   = {Source Attribution in Retrieval-Augmented Generation},
  journal = {CoRR},
  volume  = {abs/2507.04480},
  year    = {2025},
  url     = {https://arxiv.org/abs/2507.04480},
  note    = {arXiv preprint}
}

@misc{zhu2025oversight,
  title        = {Designing Meaningful Human Oversight in {AI}},
  author       = {Zhu, Liming and Lu, Qinghua and Ming, Ding and Lee, Sung Une and Wang, Chen},
  year         = {2025},
  month        = sep,
  note         = {SSRN working paper},
  doi          = {10.2139/ssrn.5501939},
  url          = {https://ssrn.com/abstract=5501939}
}

\appendix

\end{document}